\documentclass{IOS-Book-Article}

\usepackage{mathptmx}
\usepackage{float}
\usepackage{array}
\usepackage{makecell}
\usepackage{xcolor} 
\usepackage{soul}   
\usepackage{amsmath}
\usepackage{algorithm}
\usepackage{algorithmic}
\usepackage{amssymb}
\usepackage{graphicx}
\usepackage{soul}\setuldepth{article}
\def\hb{\hbox to 11.5 cm{}}

\begin{document}
\pagestyle{headings}
\def\thepage{}
\begin{frontmatter}              

\title{Image Clustering Algorithm Based on Self-Supervised Pretrained 
Models \\and Latent Feature Distribution Optimization\\ }

\markboth{}{April 2022\hb}

\author[A]{\fnms{First} \snm{Liheng Hu}}
\author[B]{\fnms{Second} \snm{Qiuyu Zhu}%
\thanks{Corresponding Author: Qiuyu Zhu, zhuqiuyu@staff.shu.edu.cn.}},
and
\author[C]{\fnms{Third} \snm{Sijin Wang}}

\runningauthor{B.P. Manager et al.}
\address[A]{School of Communication and Information Engineering, Shanghai University}
\address[B]{School of Communication and Information Engineering, Shanghai University}
\address[C]{School of Communication and Information Engineering, Shanghai University}

\begin{abstract}
In the face of complex natural images, existing deep clustering algorithms fall significantly short in terms of clustering accuracy when compared to supervised classification methods, making them less practical. This paper introduces an image clustering algorithm based on self-supervised pretrained models and latent feature distribution optimization, substantially enhancing clustering performance. It is found that : (1) For complex natural images, we effectively enhance the discriminative power of latent features by leveraging self-supervised pretrained models and their fine-tuning, resulting in improved clustering performance. (2) In the latent feature space, by searching for k-nearest neighbor images for each training sample and shortening the distance between the training sample and its nearest neighbor, the discriminative power of latent features can be further enhanced, and clustering performance can be improved. (3) In the latent feature space, reducing the distance between sample features and the nearest predefined cluster centroids can optimize the distribution of latent features, therefore further improving clustering performance. Through experiments on multiple datasets, our approach outperforms the latest clustering algorithms and achieves state-of-the-art clustering results. When the number of categories in the datasets is small, such as CIFAR-10 and STL-10, and there are significant differences between categories, our clustering algorithm has similar accuracy to supervised methods without using pretrained models, slightly lower than supervised methods using pre-trained models. The code linked algorithm is https://github.com/LihengHu/semi.
\end{abstract}

\end{frontmatter}
\markboth{April 2022\hb}{April 2022\hb}

\section{Introduction}
In recent years, deep learning-based clustering methods have become increasingly prominent, utilizing the powerful representation ability of deep learning to improve clustering performance. These methods require neural networks to learn low-dimensional representations suitable for clustering while also preserving the information and structural features of the original data. However, in the face of complex natural images, existing clustering algorithms still suffer from relatively low clustering accuracy. The primary     issue lies in the difficulty of obtaining effective discriminative latent features with limited clustering samples. Therefore, we believe that improving the effectiveness of latent feature representation, by utilizing self-supervised pretrained models trained on large-scale publicly available datasets, has the potential to yield more discriminative latent features, thus effectively enhancing clustering performance.

In this paper, addressing complex natural images, we propose an image clustering algorithm based on self-supervised pretrained models and latent feature optimization. It is based on Image Clustering Algorithm Based on Predeﬁned Evenly-Distributed Class Centroids and Composite Cosine Distance(ICBPC)\cite{ICBPC}, which relied on predefined evenly-distributed class centroids and composite cosine distance. We focus on optimizing latent feature representation and distribution, designing loss functions to enhance clustering performance. Our algorithm is trained based on a self supervised pre-trained model, using the minimum cosine distance loss function and nearest neighbor loss function to enhance algorithm performance. By incorporating self-supervised pretrained models, we utilize k-nearest neighbors obtained from the pretrained model to aid in training, significantly boosting clustering performance.

Algorithm structure is shown in Figure 1. The main contributions of this paper include:
\begin{figure}[]
\hfill
\begin{center}
\includegraphics[width=1.0\textwidth]{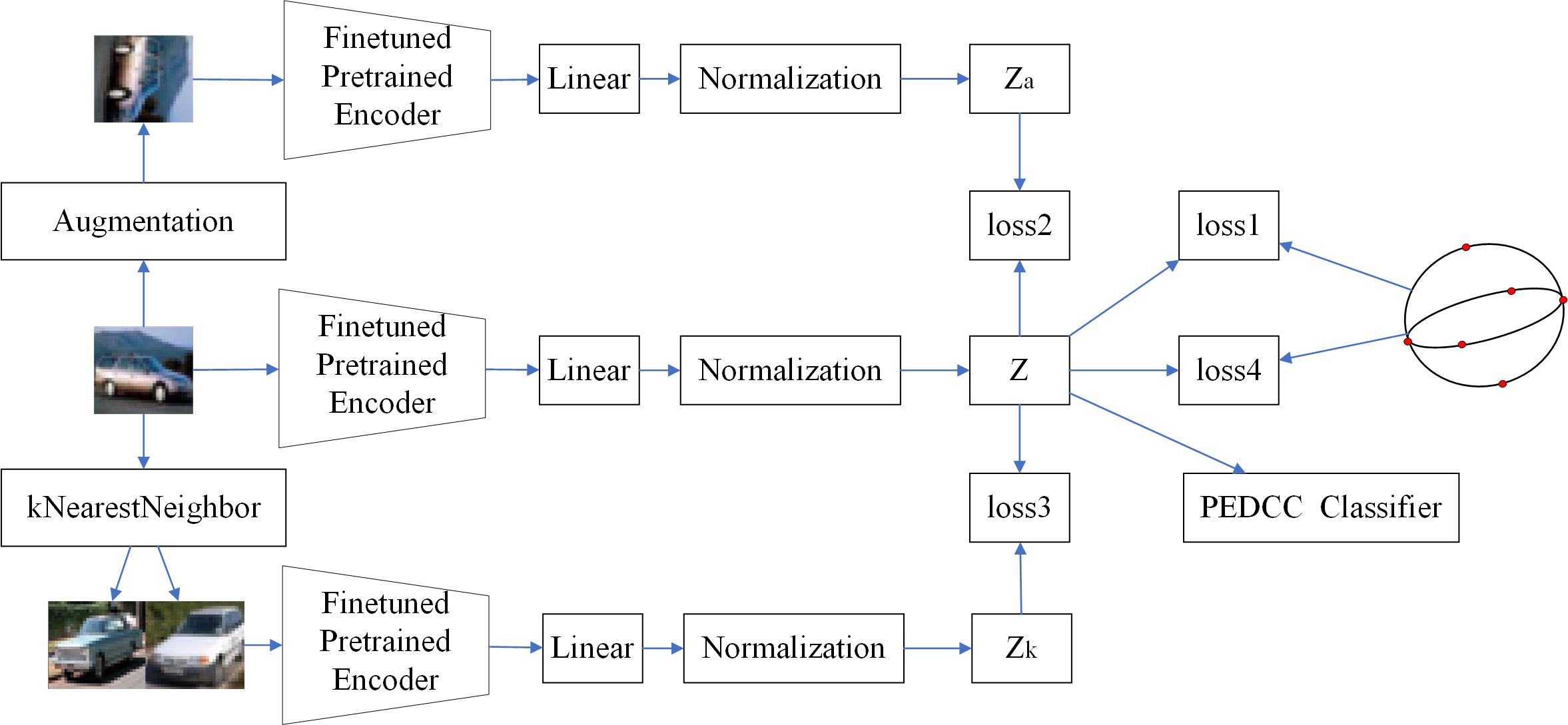}
\end{center}
\caption{
Algorithm Flow Diagram: During training, samples are channeled into the clustering network via three separate pathways. The first pathway processes the sample as-is, the second utilizes the augmented version of the sample, and the third handles the nearest-neighbor sample. Four loss functions collectively act upon the latent features outputted by the network, optimizing their representation and distribution. In the end, classification is achieved based on the minimum cosine distance between the latent features and the PEDCC points.The circle in Figure  is PEDCC.
}
\label{Fig:label}
\end{figure}

\begin{itemize} 
\item[1)]For natural image datasets, multiple self-supervised pretrained models are compared and applied to our algorithm. After further unsupervised finetuning, these models are used for subsequent clustering, enhancing the discriminative power of latent features and significantly improving clustering performance;
\item[2)]In the latent feature space, by searching for k-nearest neighbor images for each training sample and shortening the distance between the training sample and its nearest neighbor, the discriminative power of latent features can be further enhanced, and clustering performance can be improved.
\item[3)]A minimum cosine distance loss function is presented, which effectively narrows the gap between latent features and predefined evenly-distributed class centroids, resulting in reduced intra-class distances and a consequent improvement in clustering performance.
\end{itemize}
The paper is structured as follows: Section 2 encompasses a review of relevant literature, while Section 3 offers a comprehensive description of our methodology. In Section 4, we present the experimental configurations and the corresponding outcomes. Finally, Section 5 concludes the paper and provides a summary.

\section{Related Work}
This section is dedicated to presenting the development of self-supervised learning and deep clustering.

\subsection{Deep Clustering }

Clustering, a cornerstone of machine learning, frequently underpins many data mining tasks as a pivotal preprocessing stage. While conventional clustering techniques have achieved noteworthy milestones, they operate under the assumption that data instances reside in a well-structured latent vector space. However, the digital evolution of the past decades, marked by the proliferation of the internet and online services, has spurred interest in machine learning models adept at handling unstructured data devoid of explicit features, such as imagery and vast high-dimensional datasets. This renders traditional clustering methods ineffective for these data types. The dawn of representation learning within deep learning has shown promise, especially in deciphering unstructured and voluminous data. A growing body of researchers in the deep clustering domain are now leveraging deep learning paradigms to augment clustering efficacy.

Deep clustering hinges on two pivotal modules: representation learning and clustering. The former is tasked with distilling useful data representations from the raw data – a feat usually accomplished by sophisticated deep learning models. These models adeptly transform raw data into formats that unveil the data's intrinsic structures and nuances. The overarching goal of representation learning is to metamorphose raw data into a format that dovetails seamlessly with subsequent clustering endeavors.

Once a representation of the data is obtained from the representation learning module, the task of the clustering module is to group the data instances according to the similarity of these representations. This is usually achieved by applying various clustering algorithms, such as K-means\cite{kmeans}, spectral clustering\cite{kmeans}, hierarchical clustering\cite{herarchical}, etc. The goal of these algorithms is to find a way to group data instances so that instances within the same group are similar and instances in different groups are not.

Autoencoders\cite{AE1,AE2}, a mainstay in unsupervised learning, strive to encapsulate data in a compact representation, bridging input data with a lower-dimensional latent space, and subsequently reverse-engineering the input. When integrated into clustering frameworks, autoencoders pinpoint the quintessential data traits. The resulting encoded metrics then inform the clustering procedures. By channeling data through this latent space and leveraging the encoded traits, clustering efforts benefit from enhanced feature representation and discernment. Case in point: Deep Embedded Clustering(DEC)\cite{DEC} and Image Clustering Auto-Encoder(ICAE)\cite{ICAE}, both epitomizing autoencoder-centric deep clustering methodologies, albeit with nuanced operational strategies.The clustering method based on generative adversarial network\cite{gan,gan1} uses a framework composed of generator and discriminator to train generator to generate realistic data through game mode, and train discriminator to distinguish generated data from real data. The samples generated by the generator can then be used to perform the clustering task, which can achieve the clustering effect while learning the data distribution.

In deep clustering, representation learning and clustering are often conducted interactively, meaning that the clustering results can in turn guide presentation learning and vice versa. This interaction can help the model better understand the structure of the data, thereby improving the performance of the clustering. According to the interaction between representation learning and clustering, deep clustering methods are divided into two categories: multi-stage deep clustering and iterative deep clustering.

Multistage deep clustering methods first learn representations from data, and then apply traditional clustering methods to these representations. This process is carried out in different stages and is therefore called "multi-stage". However, they also have some limitations. For example, most representation learning methods are not specifically designed for clustering tasks, which may limit their ability to differentiate clusters. In addition, the clustering results cannot be further used to guide representation learning.
Such as Instance Discrimination and Feature Decorrelation(IDFD)\cite{IDFD}, this is a method of learning representations whose goal is to learn similarities between instances and reduce correlations within features. Simple K-means clustering on learned representations can also yield competitive clustering results on many existing deep clustering methods.

The method of iterative deep clustering is iteratively updated between two steps: computing the clustering result based on the current representation, and updating the representation based on the current clustering result. This allows for more interaction between presentation learning and clustering.
DeepCluster\cite{deepCluster} is a representative approach that alternates between K-means clustering and updating networks (including classifiers) by minimizing the gap between predicted cluster assignments and pseudo-labels. In fact, DeepCluster has been applied as a mature clustering algorithm in video clustering. SCAN\cite{scan} is an approach that follows the pre-training-fine-tuning framework. The clustering results are fine-tuned by self-labeling, highly confident instances are selected by thresholding soft distribution probabilities, and the entire network is updated by minimizing the cross-entropy loss of the selected instances. SPICE\cite{spice} is another representative iterative deep clustering method in which the classification model is first trained under the guidance of pseudo-labels and then semi-supervised trained on a set of reliable label instances. Our algorithm also uses this method.

\subsection{Image Clustering Algorithm Based on Predefined Evenly-Distributed Class Centroids}

The clustering algorithms discussed earlier typically compute the posterior probabilities of the samples' category membership in their clustering loss functions. They then constrain these probabilities using either Softmax or KL divergence. As a result, their impact on latent features is somewhat indirect. In contrast, our previously introduced ICAE and ICBPC approaches are entirely predicated on optimizing the distribution of latent features. By predetermining clustering points within these latent features, we've achieved improved clustering performance. This paper seeks to further refine this foundation. PEDCC represents class center points uniformly distributed on the unit hypersphere in the latent feature space. It serves as a training target for the network to maximize inter-class distance. Its three-dimensional visualization is depicted in Figure 2, where n signifies the number of class center points. The first subgraph has two predefined class centers. The number of predefined class centers increases in subsequent images.

 \begin{figure}[h]
\hfill
\begin{center}
\includegraphics[width=1.0\textwidth]{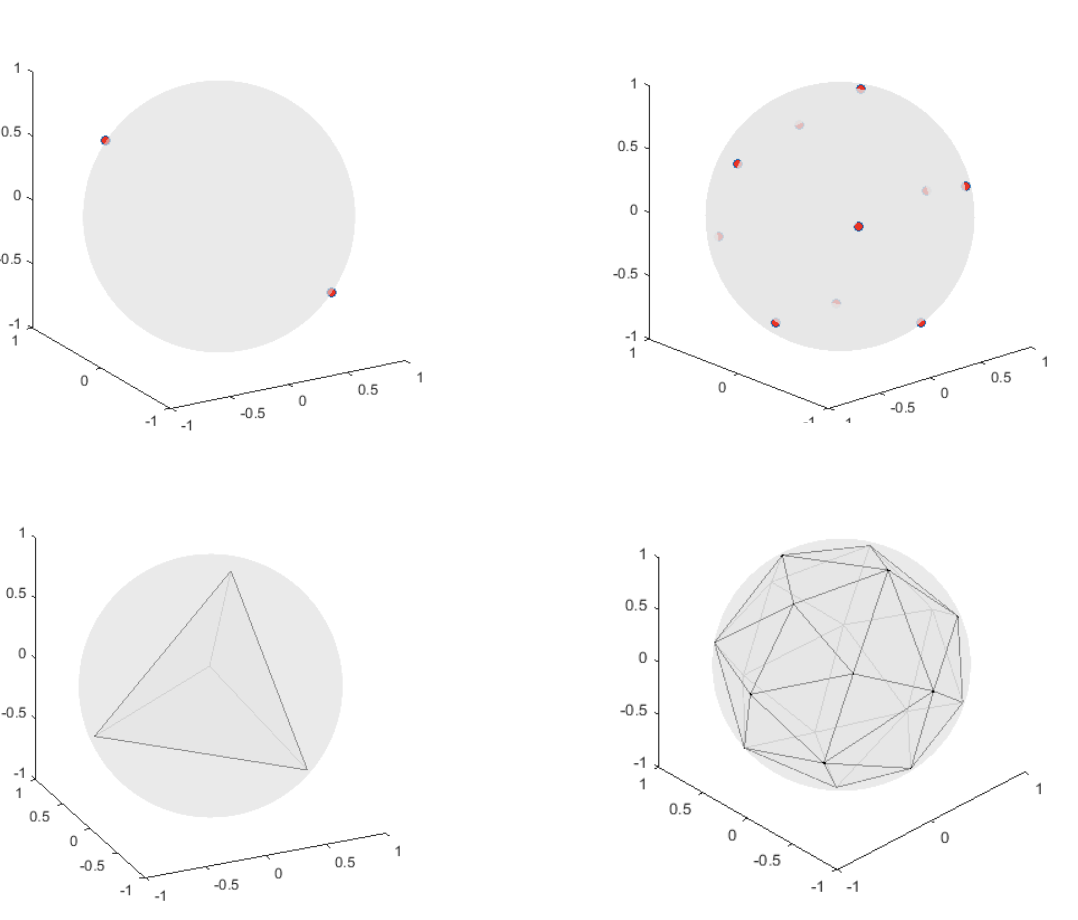}
\end{center}
\caption{Visual image of PEDCC points\cite{ICBPC}}
\label{Fig:label}
\end{figure}
ICAE merges predefined evenly-distributed class centroids with an autoencoder to yield superior outcomes. The principal distinction between the ICAE algorithm and contemporary methodologies is its architecture, loss function design, and distance metrics. While the autoencoder can produce commendable results, its intricate structure demands extensive training durations. The ICBPC algorithm simplifies this structure by solely utilizing the encoder and discarding the decoder, introducing a combined cosine distance in the process. The performance of this algorithm surpasses those harnessing autoencoders. Within the ICBPC framework that employs just the encoder's network architecture, PEDCC is utilized as the clustering center to ensure the maximal distance between latent feature classes. Constraints based on the data distribution Maximum Mean Discrepancy(MMD)\cite{mmd}, alongside contrastive constraints between samples and augmented samples, are integrated to boost clustering performance. 

Employing PEDCC and MMD can already lead to satisfactory clustering results. However, the discriminative effectiveness of the latent features is primarily constrained through contrastive learning, leaving room for optimizing the feature representation. This paper advances several tactics to bolster feature discrimination and distribution. Initially, a pretrained model is utilized, and nearest-neighbor samples for the training set are sought. A tailored loss function is then devised to narrow the gap between training samples and their nearest neighbors, enhancing feature discrimination. Subsequently, the loss function further tightens the proximity between latent features and class centers, refining the distribution of latent features.

PEDCC is generated by a mathematical model, and the exact solution is obtained by formula iterative calculation.
\begin{equation}
u_i^j=\frac{\sqrt{j(j+2)}}{j+1} w_i^j-\frac{1}{j+1} u_{j+2}^j, i=1,2, \ldots, j+1
\end{equation}

Before the generation starts, the algorithm defines n-dimensional positive and negative basic points. The coordinates of the positive fundamental points are of the form (0, 0,... , 0, 1), negative basic point coordinates are of the form (0, 0,... , 0, -1). The process of generating k uniformly distributed class centers for hyperspheres in n-dimensional space is as follows:

(1) Generate positive and negative basic points of (n-k+2) dimension;

(2) Extend the two basic points obtained in the previous step by 1 dimension and assign them to 0, so as to obtain a positive basic point on the n-dimensional space. The positive fundamental points of (n-k+2) dimension are generated, and two new points of (n-k+3) dimensional space are calculated by the formula, and then the (n-k+3) dimensional space has three evenly distributed points. The generation of these two new points is obtained by transforming the positive fundamental points. In this case, there are three evenly distributed points in the n-dimensional space;

(3) Keep repeating the second step, expanding the existing basic points by one dimension and calculating new points each time, until k evenly distributed points are obtained on the hypersphere of N-dimensional space.

\subsection{Self-supervised learning and pretrained models}

Self-supervised models can be broadly categorized into two types: generative models and discriminative models. Generative models take raw data as input, map it to a latent space, and then regenerate it back to its original form using generators. Prominent examples of such models include autoencoders and GAN (Generative Adversarial Networks). Discriminative models, on the other hand, aim to identify distinguishing features of the original data without necessarily reconstructing it at the pixel level. Initially, discriminative techniques were achieved by constructing various auxiliary tasks to facilitate self-supervised learning on unlabeled data. Commonly used auxiliary tasks were based on background and temporal sequences. However, with the rise of contrastive learning, most contemporary techniques lean toward its adoption.

Contrastive learning emphasizes discernment within the feature space, allowing it to overlook pixel-specific details in favor of abstract semantic information. This approach not only streamlines optimization compared to pixel-level reconstruction but has also witnessed significant strides in visual feature learning in recent years, rivaling or even surpassing supervised learning in certain downstream tasks, such as Momentum Contrast (MoCo)\cite{moco}.

SimCLR\cite{simclr} employs augmented data samples as positive instances, contrasting them against other samples as negatives. This method illuminated how combining multiple data augmentations is critical for crafting effective representation-defining contrastive prediction tasks. Moreover, unsupervised contrastive learning benefits from stronger data augmentation than its supervised counterpart. Introducing a learnable non-linear transformation between representations and contrastive losses has been observed to considerably enhance the quality of learned representations.

Barlow twins\cite{barlow} deviate from the norm by neither utilizing negative samples nor asymmetric structures. Instead, it introduces a novel loss function, aptly termed the "redundancy reduction loss function", to thwart model collapse.

With the advent of Masked Image Modeling (MIM) like Masked Autoencoder (MAE)\cite{mae}, a fresh self-supervised learning trend has emerged. MIM capitalizes on Vision Transformers (ViT)\cite{vit} to reconstruct masked images directly. Fundamentally, MIM is a self-supervised representation learning algorithm. Its work steps include segmenting the input image, applying random masks, and subsequently predicting attributes of the masked regions. Through MIM, encoders can attain robust representations, which in turn, foster admirable generalization in downstream tasks.

In conclusion, the essence of self-supervised learning pivots on acquiring enhanced feature representations to be deployed in subsequent tasks.

\section{Method}
In this section, we will introduce the Image Clustering Algorithm Based on Pretrained Models, the associated loss functions, the pre-trained models employed, and the network architecture.

\subsection{Image Clustering Algorithm Based on Self-Supervised Pretrained 
Models and Latent Feature Distribution Optimization(ICBPL)}
The implementation procedure of the ICBPL algorithm is presented as Algorithm 1. The algorithm takes an unlabeled image set, X, as input and outputs k image clusters. Initially, through a self-supervised pre-training model, we identify the nearest neighboring images of the latent features of the images. Concurrently, the input images undergo random image augmentation. The original images, augmented images, and nearest neighboring images are fed into the encoder, resulting in latent features for each. Three loss functions are then employed to optimize and constrain the distribution of these latent features. The MMD ensures that the feature distribution of samples closely aligns with the predefined evenly-distributed class centroids distribution. The augmentation loss function minimizes the distance between the original and augmented samples, while the K-nearest neighbors loss function reduces the distance between K-nearest samples. The minimal cosine distance loss function narrows the distance between sample features and predefined evenly-distributed class centroids, shrinking intra-class distances and expanding inter-class gaps. Throughout the training process, we also periodically update the K-nearest neighbors using the most recent network to ensure precision in identifying the nearest neighboring images.The stopping condition of the algorithm is that ACC reaches its maximum value and no longer increases.In the 5.4 experiment, we conclude that the best clustering performance can be obtained by updating every 30 epochs.Figure 3 shows a schematic diagram of the algorithm.

\begin{algorithm}[H]
		\caption{ICBPL~algorithm}\label{1234}
		\label{alg::conjugateGradient}
		\begin{algorithmic}[1]
\REQUIRE
    $X$ = unlabeled images;
    \ENSURE
      $K$ classes of clustering images;
    \STATE Finetuning Barlow Twins pretrained model
    \STATE Initialize PEDCC cluster centers;  
    \REPEAT  
    \STATE if epoch mod 30 = 0 , update k nearest neighbor 
      \STATE  $\widehat{X_a}$=Augumentation($X$); $\widehat{X_k}$=kNearestNeighbor($X$);  
      \STATE  $\widehat{Z_a}$ = Encoder($\widehat{X_a}$);
       $\widehat{Z_k}$ = Encoder($\widehat{X_k}$);
       
       $Z$ = Encoder($X$);  
      \STATE  
        $loss 1$ = MMD($\mathrm{Z}$, PEDCC);

    $loss 2$ = Contrastive loss($\mathrm{Z}$, $\widehat{\mathrm{Z_a}}$);

    $loss 3$ =  Contrastive loss($\mathrm{Z}$, $\widehat{\mathrm{Z_k}}$)
    ; 
      
    $loss 4$ = MinCosDistance($\mathrm{Z}$,PEDCC);
    \UNTIL Stopping criterion meet  
		\end{algorithmic}
	\end{algorithm}

  \begin{figure}[h]
\hfill
\begin{center}
\includegraphics[width=0.3\textwidth]{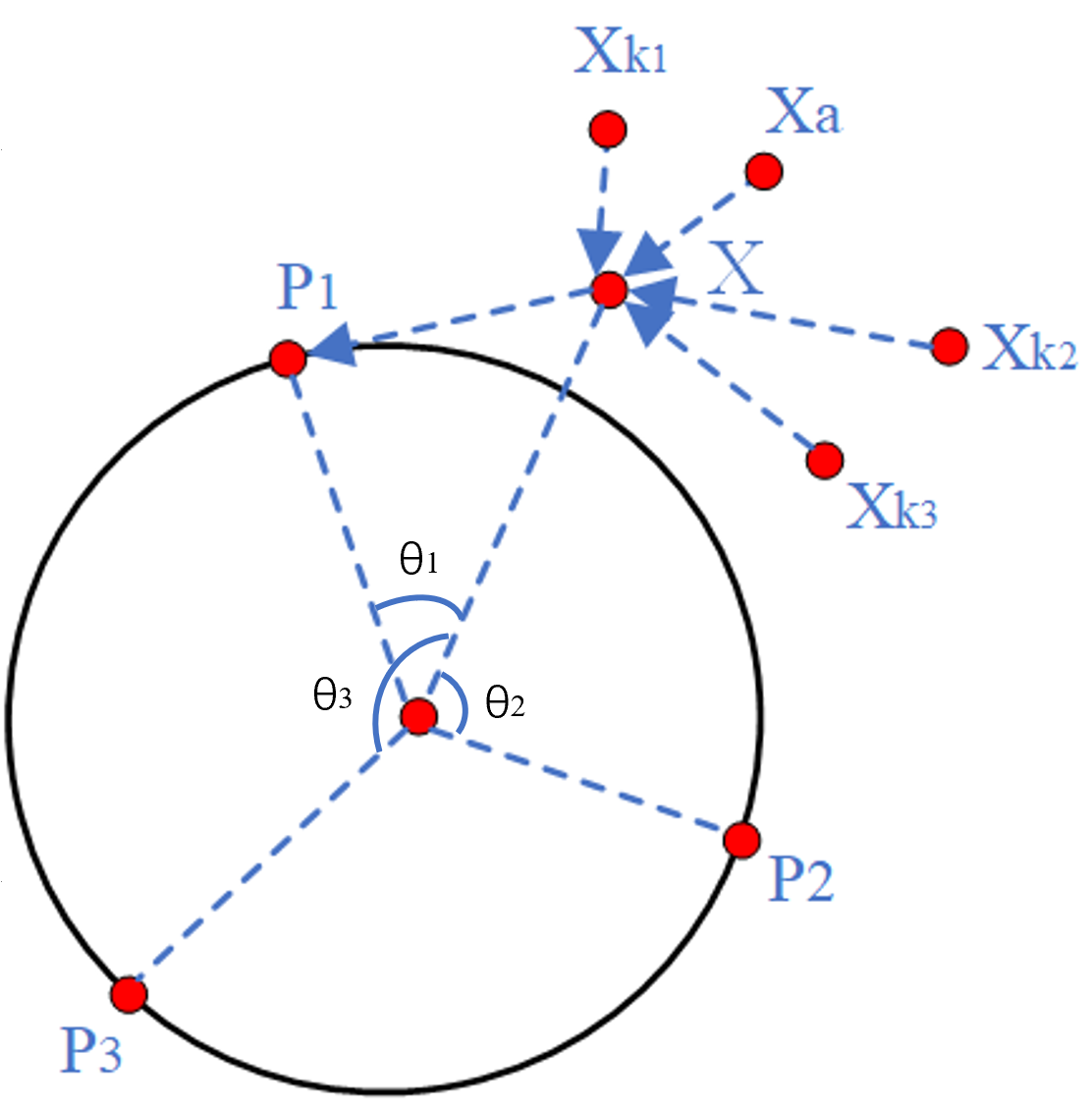}
\end{center}
\caption{Algorithm diagram , Where $P$ is the pedcc point, $X$ is the original sample, $X_k$ is the nearest neighbor sample, and $X_a$ is the augmented sample.}
\label{Fig:label}
\end{figure}

\subsection{Loss function}
In this paper, an image clustering algorithm based on pretrained model is proposed. It is an improvement on the image clustering algorithm proposed by us based on pre-defined evenly distributed class centroid and complex cosine distance. Based on ICBPC, we add the minimum cosine distance loss function and the K-nearest neighbor loss function, and apply them to the self-supervised pre-training model.
There are four loss functions in this algorithm. Loss1 and Loss2 use the same loss function as the ICBPC algorithm. Loss1 is the maximum mean discrepancy loss function, and the Loss1 formula is as follows:
$${loss1 = MMD([Z,\widehat{Z_a}], PEDDC)}$$
$${=\frac{1}{M(M-1)} \sum_{i \neq j}^{M} k\left(l_{i}, l_{j}\right)+\frac{1}{c(c-1)} \sum_{i \neq j}^{C} k\left(u_{i}, u_{j}\right)}$$
\begin{equation}-\frac{2}{M C} \sum_{i, j=1}^{M, C} k\left(l_{i}, u_{j}\right).\label{eq}\end{equation}
where $Z$ is the intermediate latent features, $\widehat{Z}$ mean the latent features of the augmented data, $M$ mean its dimension, $l_i=[Z,\widehat{Z}]$  is the latent features of the image and its augmented latent features; $u_i$ represents the PEDCC class centers, $C$ is its number, $k(x,y)$ is  kernel  function.The formula for the kernel function is as follows:
\begin{equation}
K({x}_1, {x}_2) = \exp\left(-\frac{\|{x}_1 - {x}_2\|^2}{2\sigma^2}\right)
\end{equation}

Loss2 is the augmentation loss function based on the cosine distance. The Loss2 formula is as follows:

\begin{equation}loss_{2}(\textbf{z},\textbf{z}_a)=\frac{1}{2N} \sum_{i=0}^{N} d^2 = \frac{1}{2N} \sum_{i=0}^{N} (1-z_{i}*z_{a_i})^2 \label{eq}
\end{equation}
where $d$ represents the cosine distance between two samples, 
$z$  stands for the original
sample, and $z_a$  signifies the augmented sample. The primary purpose of data augmentation is to mitigate overfitting in the network and aid in extracting more discriminative features. By varying the training images, one can obtain a network with enhanced generalization capabilities, making it better suited for practical applications. In our experiments, we employed random cropping and resizing, random horizontal flipping, color jittering, random grayscale conversion, and Gaussian blurring. Employing a combination of data augmentation techniques yields superior generalization performance.

Loss3 is a K nearest neighbor loss function based on cosine distance. The Loss3 formula is as follows:
\begin{equation}loss_{3}(\textbf{z},\textbf{z}_k)=\frac{1}{2NM} \sum_{i=0}^{N}\sum_{j=0}^{M} d^2  = \frac{1}{2NM} \sum_{i=0}^{N}\sum_{j=0}^{M} (1-z_{i}*z_{k_j})^2 \label{eq}
\end{equation}
where $d$ represents the cosine distance between the two samples, 
$z_i$ represents the feature of original sample,and $z_k$ represents the feature of K nearest neighbor sample.

Loss4 is the minimum cosine distance loss function.
The minimum cosine distance loss function is to further narrow the distance between the feature and the class center and reduce the distance within the class. The formula is as follows:
\begin{equation}{loss_{4} = (1-\max cos\theta_{p_i})^2 \label{eq}}\end{equation}
where $\theta_{p_i}$ stands for the angle between the latent feature and the PEDCC point , 
as shown in Figure 3. By leveraging the minimum cosine distance function to identify the PEDCC point nearest to the latent feature and reduce their distance, clustering performance can be effectively enhanced.

Integrating the trio of the previously detailed loss functions, the composite loss function emerges as:
\begin{equation}loss= {loss}_{1}+ \lambda_{1} \times {loss}_{2} + \lambda_{2} \times {loss}_{3} + \lambda_{3} \times {loss}_{4}.\label{eq}\end{equation}
In this configuration, 
 $\lambda$ operates as the modulating weight for every individual loss function. For different datasets, the weights assigned to the four loss functions will be adjusted, and different weightings will lead to varied outcomes. The specific weightings are detailed in Table 1. Optimal clustering performance is achieved when weights align with those presented in the table.

\begin{table}[h]
\caption{Loss function weight.}
\begin{center}
\begin{tabular}{|l|l|l|l|}
\hline
Datasets & $\lambda_{1}$   &  $\lambda_{2}$ & $\lambda_{3}$    \\ \hline
CIFAR-10   & 9 & 2 & 2  \\ \hline
STL-10   & 8 & 2 & 2  \\ \hline
CIFAR-100   & 8 & 2 & 2  \\ \hline
ImageNet-50   & 8 & 2 & 2  \\ \hline
\end{tabular}
\end{center}
\end{table}
For the classification of samples, we employ the minimum cosine distance between the latent features and the PEDCC points for categorization, as follows:
\begin{equation}
J = \arg\max_{i} \left[ d(Z, \text{PEDCC}_i) \right]
\end{equation}
where $J$ stands for the category number after classification , 
$d$ signifies the cosine distance, and $PEDCC_{i}$ is the $i^{th}$ PEDCC point.

\subsection{Using Self-supervised pretrained models }
Self-supervised pre-trained models refer to those trained via self-supervised learning techniques. These methodologies don't necessitate human-annotated data but learn feature representations by training on vast volumes of raw data. Suitable for tasks spanning from natural language processing to computer vision, these pre-trained models can act as foundational models for further training endeavors. Widely recognized pre-trained models in this domain include SimCLR, Barlow twins, and MAE.

Traditional clustering algorithms often grapple with a persistent challenge: extracting highly discriminative latent features from a limited set of clustering samples. To address this bottleneck, we propose a solution that harnesses self-supervised pre-trained models, trained on expansive public datasets. This strategy more adeptly extracts highly discriminative latent features, thereby amplifying the efficiency and performance of clustering algorithms.

In our clustering tasks, we employ pre-trained models of the aforementioned three self-supervised algorithms, all trained on the ImageNet dataset. Here, deep clustering operates as a downstream task of self-supervised learning. These pre-trained models fortify our clustering algorithm by furnishing richer and more effective feature representations, thereby enhancing the algorithm's clustering ability, especially when confronted with intricate natural images.

\subsection{Training strategy}
Within the realm of machine learning, fine-tuning refers to the practice of adjusting a pre-trained model to enhance its prediction capabilities on a new dataset. This method has gained traction, as training a high-quality model demands substantial data, and harnessing a pre-trained model can curtail both time and computational overheads. Typically, during the fine-tuning process, the majority of the parameters from the pre-trained model are retained, with the primary focus being on the last few layers to better align with the new dataset. An inherent advantage of this method is that the pre-trained model has already assimilated myriad useful features, thus providing a robust starting point for the new model.

The training strategy outlined in this article emphasizes the fine-tuning of the network, notably by freezing several convolutional layers from the pre-trained model. It's often the majority of layers closer to the input that are frozen since they encapsulate a wealth of foundational information. The layers that undergo training usually encompass convolutional layers closer to the output and the fully connected layers.

Self-supervised pre-trained models can significantly bolster the performance of clustering algorithms, and a judicious training strategy can optimize the results. Predominantly, this paper employs two tactics: 1. Training only the last two convolutional layers and 2. Training the entire network.

Concurrently, throughout the training phase, we refresh the k-nearest neighbors for the original samples to procure a more accurate set of k-nearest neighbors, enhancing clustering precision. A pertinent aspect to address is the timing of these updates. Subsequent experiments have discerned that dynamically updating the k-nearest neighbors every 30 steps yields superior clustering performance.

\section{Experiments and Discussions}
\subsection{Experiments Settings}
\subsubsection{Datasets}
We evaluated the performance of our algorithm using four natural image datasets. These datasets are CIFAR-10, STL-10, CIFAR-100, and ImageNet-50, as shown in Table 2. For the ImageNet-50 dataset, we randomly selected images from 50 categories within the ImageNet dataset. Before feeding them into the network, all datasets were normalized to the range [-1, 1]. To accommodate the pre-trained models, all samples were resized to a resolution of 224x224 before clustering.
\begin{table}[h]

\caption{Datasets}
\centering
\begin{tabular}{|l|l|l|l|}
\hline
\textbf{Datasets}	& \textbf{Samples}	& \textbf{Categories}& \textbf{Image Size}	\\ \hline
CIFAR-10  & 60,000   & 10      & 32 ${\times}$ 32 ${\times}$ 3         \\ \hline
STL-10  & 5000   & 10      & 96 ${\times}$ 96 ${\times}$ 3         \\ \hline
CIFAR-100-20  & 60,000   & 20      & 32 ${\times}$ 32 ${\times}$ 3         \\ \hline
ImageNet-50 &50000 & 50 & 224 ${\times}$ 224 ${\times}$ 3         \\ \hline
\end{tabular}
\end{table}

\subsubsection{Experimental Setup}
Before initiating the experiments, we established the number of classification categories and the dimensions of the intermediate layer features. The initial learning rate was set at 0.001, utilizing the Adam optimizer. We defined the batch size as 100 and the training iterations as 400. Throughout the training, the network architecture remained consistent. Hyperparameter configurations are detailed in Table 1. The values in Table 1 represent the settings that yielded optimal clustering results. All experimental outcomes are averaged following four rounds of training.

 \subsubsection{Evaluation Metrics}
The following two indicators are used to validate our algorithm: Cluster Accuracy (ACC) \cite{ACCNMI}, and Normalized Mutual Information (NMI) \cite{ACCNMI}. 

Clustering accuracy is an intuitive measure of a clustering algorithm's ability to correctly assign samples from a dataset to their respective categories. The formula is:
\begin{equation}
A C=\frac{\sum_{i=1}^n \delta\left(s_i, \operatorname{map}\left(r_i\right)\right)}{n}
\end{equation}
where $r_i$ is the obtained label corresponding to the sample,  $s_i$ is the real label of the sample, $n$ is the total number of samples, and $\delta$ represents the function as follows:
\begin{equation}
\delta(x, y)= \begin{cases}1 & \text { if } x=y \\ 0 & \text { otherwise }\end{cases}
\end{equation}

NMI stands for normalized mutual information. NMI is a metric in information theory used to assess the degree of similarity between the results of two clusters. It is based on mutual information in information theory, and it is standardized. The calculation of NMI requires the distribution of real labels and clustering results to quantify the similarity between them. NMI values typically range from 0 to 1, with 1 indicating that the two clustering results are exactly the same and 0 indicating that there is no similarity between them. The formula is:
\begin{equation}
N M I(\Omega ; C)=\frac{I(\Omega ; C)}{\frac{H(\Omega)+H(C)}{2}}
\end{equation}

In this formula, $I$ represents mutual information and $H$ represents entropy.

\subsection{Analysis on Computational Time}
Our proposed algorithm simplifies the algorithm structure by using only the encoder and discarding the decoder. The proposed algorithm is compared with the ICAE algorithm using autoencoder. As shown in the Table 3, the time per epoch of ICBPL is shorter. Thus, ICBPL algorithm has a lower computational complexity.

\begin{table}[H] 
\caption{Computational complexity comparison}
\centering
\begin{tabular}{|l|l|l|l|}
\hline
\textbf{Datasets}	& \textbf{Encoder-only}	& \textbf{Auto-encoder}& \textbf{Training time of each epoch (seconds)} 	\\ \hline
CIFAR-10  & \checkmark  & & 98    \\ 
CIFAR-10  & &\checkmark    & 132  \\        \hline
\end{tabular}
\end{table}

\subsection{Ablation Experiment}
We conducted ablation studies to evaluate the clustering performance based on the configuration of the number of k-nearest neighbor samples as well as to assess the efficacy of each loss function.
\subsubsection{The influence of the number of K-nearest neighbor samples}
The clustering accuracy varies with different settings for the number of k-nearest neighbor samples. We validated this assertion on the STL-10 dataset. As presented in Table 4, it can be observed that setting the number of k-nearest neighbors to 4 yields the best clustering performance.

\begin{table}[h]
\caption{Effectiveness of the number of K-nearest neighbor samples on clustering results}
\begin{tabular}{|l|l|}
    \hline
    \textbf{K-nearest neighbor samples} & \textbf{ACC} \\ \hline
    0 & 0.608 \\ 
    1 & 0.896 \\ 
    2 & 0.897 \\
    3 & 0.901 \\
    4 & \bfseries0.905 \\
    5 & 0.904 \\ \hline
  \end{tabular}
\end{table}

\subsubsection{Effectiveness of different distance measures}
Within the framework of our contrastive and k-nearest neighbor loss functions, we have chosen cosine distance as our distance measure. When juxtaposed with the Euclidean distance, cosine distance emerges as a more fitting metric for our model and methodology. We performed validation on the STL-10 dataset. Table 5 reveals a marked improvement in performance when leveraging cosine distance.

\begin{table}[h]
\caption{Influence of the number of K-nearest neighbor samples on clustering results}
\centering
\begin{tabular}{|l|l|l|}
\hline
 	& \textbf{Euclidean distance}	& \textbf{Cosine distance} 	\\ \hline
\textbf{ ACC}	& 0.896	& \bfseries0.905	\\ \hline

\end{tabular}
\end{table}

\subsubsection{Effectiveness of different loss functions}
The effectiveness of each loss function was validated through experiments, with the results presented in Table 6. The pre-trained models used for the data were those that yielded the best clustering performance for each dataset. The experimental outcomes indicate that the combination of all four loss functions provides the most optimal clustering results.

\begin{table}[h]
\caption{Ablation Experiment Results}
\centering
\scalebox{1}{
\begin{tabular}{|c|c|c|c|c|c|c|}
\hline
Datasets & Loss1 & Loss2 & Loss3 & Loss4 & ACC & NMI \\
\hline
CIFAR-10 & \checkmark & - & - & - & 0.124 & 0.113 \\
CIFAR-10 & \checkmark & \checkmark & - & - & 0.785 & 0.754 \\
CIFAR-10 & \checkmark & \checkmark & \checkmark & \checkmark & \bfseries0.895 & \bfseries0.813 \\
CIFAR-100 & \checkmark & - & - & - & 0.09 & 0.08 \\
CIFAR-100 & \checkmark & \checkmark & - & - & 0.432 & 0.406 \\
CIFAR-100 & \checkmark & \checkmark & \checkmark & \checkmark & \bfseries0.527 & \bfseries0.502 \\
STL-10 & \checkmark & - & - & - & 0.186 & 0.157 \\
STL-10 & \checkmark & \checkmark & - & - & 0.798 & 0.646 \\
STL-10 & \checkmark & \checkmark & \checkmark & \checkmark & \bfseries0.905 & \bfseries0.753 \\
ImageNet-50 & \checkmark & - & - & - & 0.11 & 0.10 \\
ImageNet-50 & \checkmark & \checkmark & - & - & 0.682 & 0.765 \\
ImageNet-50 & \checkmark & \checkmark & \checkmark & \checkmark & \bfseries0.770 & \bfseries0.831 \\
\hline
\end{tabular}
}
\end{table}

\subsection{Effectiveness of different pretrained models}
In our experiments, for datasets CIFAR-10, CIFAR-100, STL-10, and ImageNet-50, we utilized three self-supervised pre-trained models—SimCLR, Barlow twins, and MAE—all originally trained on the ImageNet dataset. The goal was to determine the impact of these pre-trained models on clustering performance. Table 7 captures the clustering results for each dataset using the various pre-training models.

\begin{table}
  \centering
  \caption{Clustering Performance of Different Pretrained Models}
\scalebox{1.0}{
  \begin{tabular}{|c|c|c|c|c|c|c|c|c|}
    \hline
              & \multicolumn{2}{c|}{CIFAR-10} & \multicolumn{2}{c|}{STL-10} & \multicolumn{2}{c|}{CIFAR-100} & \multicolumn{2}{c|}{ImageNet-50} \\
    \hline
    Model & NMI & ACC & NMI & ACC & NMI & ACC & NMI & ACC \\
    \hline
    No Model & 0.125 & 0.241 & 0.205 & 0.293 & 0.071 & 0.082 & 0.063 & 0.068 \\
    \hline
    SimCLR & 0.805 & 0.843 & 0.734 & 0.885 & 0.482 & 0.487 & 0.821 & 0.743 \\
    \hline
    Barlow Twins & \bfseries0.813 & \bfseries0.895 & \bfseries0.753 & \bfseries0.905 & \bfseries0.502 & \bfseries0.527 & \bfseries0.831 & \bfseries0.770 \\
    \hline
    MAE & 0.701 & 0.802 & 0.687 & 0.821 & 0.442 & 0.457 & 0.751 & 0.702 \\
    \hline
  \end{tabular}
}
\end{table}

The clustering performance of SimCLR and Barlow twins is close, and the performance of Barlow twins is the best. However, the clustering performance obtained by MAE is inferior to the other two pre-trained models. It can be seen that the pre-trained model based on Resnet network is more suitable for our image clustering algorithm. The pre-training method of contrast learning is similar to the latent feature ideas used in deep clustering, so the pretrained model helps to obtain more discriminating features.

\subsection{Effectiveness of of different training strategies}
The potential of self-supervised pre-trained models in elevating the performance of clustering algorithms is remarkable, and adopting the right training strategy can yield optimal outcomes. For our approach, we have chosen the Barlow twins pre-trained model, which aligns best with our algorithm's requirements. Before embarking on the training phase, we fine-tune our network with the Barlow Twins self-supervised methodology. The Barlow twins utilize a Resnet50 network for pre-training, with its architectural details outlined in Table 8.
\begin{table}[h]
\caption{ResNet-50 Network Architecture}
\centering
\scalebox{1.0}{
\begin{tabular}{|c|c|c|}
\hline
Layer & Output Size & Remarks \\
\hline
Conv1 & 112x112 & 32 channels \\
Conv2 & 56x56 & 256 channels \\
Conv3 & 28x28 & 512 channels \\
Conv4 & 14x14 & 1024 channels \\
Conv5 & 7x7 & 2048 channels, Encoder output \\
Fully connected layer & - & Latent feature \\
\hline
\end{tabular}
}
\end{table}
Table 9 illustrates the clustering performance under various training configurations. The training settings should be adapted based on the dataset in question. For STL-10, fine-tuning the two convolutional layers closest to the output yields the best results. In contrast, for CIFAR-10, CIFAR-100, and ImageNet-50, training the entire network delivers the optimal performance.

\begin{table}[h]
\caption{Clustering Performance of Different Network Architectures}
\centering
\scalebox{1.0}{
\begin{tabular}{|c|c|c|c|c|c|c|c|c|}
\hline
Training Setting & \multicolumn{2}{c|}{CIFAR-10} & \multicolumn{2}{c|}{STL-10} & \multicolumn{2}{c|}{CIFAR-100} & \multicolumn{2}{c|}{ImageNet-50} \\
\hline
\textbf{Model} & NMI & ACC & NMI & ACC & NMI & ACC & NMI & ACC \\
\hline
\textbf{\thead{Barlow Twins \\ (Last Two \\ Conv Layers)}}
& 0.783 & 0.821 & \bfseries0.753 & \bfseries0.905 & 0.476 & 0.489 & 0.802 & 0.745 \\
\hline
\textbf{\thead{Barlow Twins \\ (All Layers)}}  & \bfseries 0.813 & \bfseries0.895 & 0.741 & 0.895 & \bfseries0.502 & \bfseries0.527 & \bfseries 0.831 & \bfseries0.770 \\
\hline
\end{tabular}
}
\end{table}

The performance metrics of STL-10 under varying frequencies of dynamic k-nearest neighbor updates during training are presented in Table 10. The findings indicate that a periodic update every 30 epochs strikes the balance for achieving the pinnacle of clustering accuracy.

\begin{table}[h]
\caption{Clustering Performance with Dynamic K-nearest Neighbor Settings}
\centering
\scalebox{1.0}{
\begin{tabular}{|c|c|c|c|c|}
\hline
Number of Training Iterations & 10 & 20 & 30 & 40 \\
\hline
STL-10 & 0.891 & 0.895 & \bfseries0.906 & 0.898 \\
\hline
\end{tabular}
}
\end{table}

\subsection{Compared with the Latest Clustering Algorithm}
We compared our ICBPL clustering algorithm with the latest clustering algorithms. Our approach achieved commendable results across four datasets, demonstrating a significant enhancement in metrics. When the number of categories is relatively low (e.g., cifar-10, stl-10) and there is a marked distinction between categories, the accuracy of our clustering approach closely approximates the accuracy of supervised methods that do not utilize pre-trained models, only slightly lower than the accuracy achieved by supervised training with pre-trained models. The comparison is detailed in Table 11. In Table 11, all results are reported either by running the published code from their respective works or obtained directly from the corresponding papers. The symbol "-" indicates that the result was not applicable to the respective paper or code. Bold figures in Table 11 signify the best results.

\begin{table}[h]
\caption{Comparison with Recent Clustering Algorithm Results}
\centering
\scalebox{0.9}{
\begin{tabular}{|c|c|c|c|c|c|c|c|c|}
\hline
 & \multicolumn{2}{c|}{CIFAR-10} & \multicolumn{2}{c|}{STL-10} & \multicolumn{2}{c|}{CIFAR-100-20} & \multicolumn{2}{c|}{ImageNet-50} \\
\hline
Algorithm & NMI & ACC & NMI & ACC & NMI & ACC & NMI & ACC \\
\hline
K-means\cite{kmeans} & 0.064 & 0.199 & 0.125 & 0.192 & 0.084 & 0.130 & - & - \\
NMF-LP\cite{NMF-LP} & 0.051 & 0.180 & - & - & - & - & - & - \\
DEC\cite{DEC} & 0.057 & 0.208 & 0.276 & 0.359 & 0.136 & 0.185 & - & - \\
JULE\cite{JULE} & - & - & 0.182 & 0.277 & 0.103 & 0.137 & - & - \\
DAC\cite{DAC} & 0.396 & 0.522 & 0.366 & 0.469 & 0.185 & 0.238 & - & - \\
ADC\cite{ADC} & - & 0.293 & - & 0.530 & - & 0.160 & - & - \\
IDFD\cite{IDFD} & 0.711 & 0.815 & 0.643 & 0.756 & - & - & - & - \\
ICAE\cite{ICAE} & 0.080 & 0.215 & - & - & - & - & - & - \\
ICBPC\cite{ICBPC} & 0.182 & 0.298 & 0.525 & 0.551 & - & - & 0.375 & 0.363 \\
SCAN\cite{scan} & 0.797 & 0.883 & 0.698 & 0.809 & 0.486 & 0.507 & 0.822 & 0.768 \\
ICBPL & \bfseries0.813 & \bfseries0.895 & \bfseries0.753 & \bfseries0.905 & \bfseries0.502 & \bfseries0.527 & \bfseries0.831 & \bfseries0.770 \\
Supervised & 0.862 & 0.938 & 0.659 & 0.806 & 0.800 & 0.680 & - & - \\
Supervised on pretrained model & 0.878 & 0.956 & 0.810 & 0.926 & 0.824 & 0.720 & - & - \\
\hline
\end{tabular}
}
\end{table}

\section{Conclusion}
In this study, we introduce an image clustering algorithm that harnesses the power of self-supervised pretrainde models and optimizes the distribution of latent features.    The algorithm emphasizes the optimization of latent feature representations and their distributions. We've incorporated the minimum cosine distance loss function and the k-nearest neighbor contrast loss function, further employing them within self-supervised pre-training models.    After fine-tuning these models, the utilization of k-nearest neighbors significantly elevates the clustering performance, especially achieving near-supervised learning classification results on datasets like Cifar10 and STL-10.   
The algorithms in this paper are all tested on data sets, not to the actual application scenarios. In future work, our focus will shift to exploring pre-trained models to enhance the performance of clustering algorithms in scenarios with a large number of categories and finer granularity.

\end{document}